\documentclass[runningheads]{llncs}

\usepackage{graphicx}
\newcommand\Tau{\mathcal{T}}
\usepackage{amssymb,amsmath}
\usepackage[linesnumbered, ruled,vlined]{algorithm2e}

\newlength\mylen
\newcommand\myinput[1]{%
  \settowidth\mylen{\KwIn{}}%
  \setlength\hangindent{\mylen}%
  \hspace*{\mylen}#1\\}
  
\usepackage{tikz}
\newcommand*\circled[1]{\tikz[baseline=(char.base)]{
            \node[shape=circle,draw,inner sep=0.5pt] (char) {#1};}}
            
\usepackage{tabularx}
\usepackage{booktabs}
\usepackage{subfig}

\begin{document}
\title{Task Attended Meta-Learning for Few-Shot Learning}
\titlerunning{Task Attended Meta-Learning for Few-Shot Learning}

\author{Aroof Aimen\inst{1} \and
Sahil Sidheekh\inst{1} \and
Narayanan C. Krishnan \inst{1}}
\institute{
Indian Institute of Technology, Ropar
\\
\email{\{2018csz0001, 2017csb1104, 2017csb1119, ckn\}@iitrpr.ac.in}}

\maketitle              
\begin{abstract}

Meta-learning (ML) has emerged as a promising direction in learning models under constrained resource settings like few-shot learning. The popular approaches for ML either learn a generalizable initial model or a generic parametric optimizer through episodic training. The former approaches leverage the knowledge from a batch of tasks to learn an optimal prior. In this work, we study the importance of a batch for ML. Specifically, we first incorporate a batch episodic training regimen to improve the learning of the generic parametric optimizer. We also hypothesize that the common assumption in batch episodic training that each task in a batch has an equal contribution to learning an optimal meta-model need not be true. We propose to weight the tasks in a batch according to their ``importance" in improving the meta-model's learning. To this end, we introduce a training curriculum motivated by selective focus in humans, called task attended meta-training, to weight the tasks in a batch. Task attention is a standalone module that can be integrated with any batch episodic training regimen. The comparisons of the models with their non-task-attended counterparts on complex datasets like miniImageNet and tieredImageNet validate its effectiveness.

\end{abstract}

\section{Introduction}
The ability to infer knowledge and discover complex representations from data has made deep learning models widely popular in the machine learning community. However, these models are data-hungry, often requiring large volumes of labeled data for training. Collection and annotation of such large amounts of training data may not be feasible for many real life applications, especially in domains that are inherently data constrained, like medical and satellite image classification, drug toxicity estimation, etc. Meta-learning (ML) has emerged as a promising direction for learning models in such settings, where only a limited amount (few-shots) of labeled training data is available. A typical ML algorithm employs an episodic training regimen that differs from the training procedure of conventional learning tasks. This episodic meta-training regimen is backed by the assumption that a machine learning model quickly generalizes to novel unseen data with minimal fine-tuning when trained and tested under similar circumstances \cite{vinyals2016matching}. To facilitate such a generalization capacity, a meta-training phase is undertaken, where the model is trained to optimize its performance on several homogeneous tasks/episodes randomly sampled from a dataset. Each episode or task is a learning problem in itself. In the few-shot setting each task is a classification problem, a collection of $K$ support (train) and $Q$ query (test) samples corresponding to each of the $N$ classes. Task-specific knowledge is learned using the support data, and meta-knowledge across the tasks is learned using query samples, which essentially encodes “how to learn a new task effectively.” The learned meta-knowledge is generic and agnostic to tasks from the same distribution. It is typically characterized in two different forms - either as an optimal initialization for the machine learning model or a learned parametric optimizer.  Under the optimal initialization view, the learned meta-knowledge represents an optimal prior over the model parameters, that it is equidistant,  but close to the optimal parameters for all individual tasks. This enables the model to rapidly adapt to unseen tasks from the same distribution \cite{DBLP:conf/icml/FinnAL17,DBLP:journals/corr/LiZCL17,DBLP:conf/cvpr/JamalQ19}. Under the parametric optimizer view, meta-knowledge pertaining to the traversal of the loss surface of individual tasks is learned by the meta-optimizer. Through learning task specific and task agnostic characteristics of the loss surface, a parametric optimizer can thus effectively guide the base model to traverse the loss surface and achieve superior performance on unseen tasks from the same distribution \cite{DBLP:conf/iclr/RaviL17}.

Initialization based ML approaches accumulate the meta-knowledge by simultaneously optimizing over a batch of tasks. On the other hand, a parametric optimizer sequentially accumulates meta-knowledge across individual tasks. The sequential accumulation process leads to a long oscillatory optimization trajectory and  a bias towards the last task, limiting the parametric optimizer's task agnostic potential. Leveraging common knowledge from a batch of tasks to learn the parametric optimizer can help address this issue. We first propose to accumulate meta-knowledge in a batch mode for the parametric optimizer. Further, under such batch episodic training, a common assumption in ML that the randomly sampled episodes of a batch contribute equally to improving the learned meta-knowledge need not hold good. Due to latent properties of the sampled tasks in a batch and the model configuration, some tasks may be better aligned with the optimal meta-knowledge than others. We hypothesize that proportioning the contribution of a task as per its alignment towards the optimal meta-knowledge can improve the meta-model's learning. This is analogous to classical machine learning algorithms like bootstrapping, where samples leading to false positives are prioritized and therefore replayed. However, in this example, the latent properties due to which a sample is prioritized are explicitly defined. For complex task distributions, explicitly handcrafting the notion of ``importance" of a task would be hard. Instead, we propose a  task attended meta-training curriculum motivated by the idea of selective focus in human beings to learn the ``importance" of tasks according to their ability to improve the meta-knowledge.

Selective focus is an essential aspect of human learning. All experiences do not contribute equally to the overall learning of a human being. Some experiences have a significant impact on the human learning process than others. Consider an example of a student preparing for an examination; as memory, and time are limited resources, the student selectively focuses on topics that have the possibility of maximizing the payoff (grades in this case) and filters out the less important topics. This selective attention can maximize the payoff under the constrained setting, which, however, largely depends on the students' capability of predicting the important topics and harmonizing with the topics that actually appear in the examination. The student chooses the important topic based on some meta-information like recurrence pattern of the exact or similar topic in previous years, the complexity of the topic with respect to the scope of examination, the re-usability of a learned topic elsewhere, etc. Besides these factors, some latent properties like the question paper setter's expertise may bring some stochasticity in maximizing the payoff. 

Analogously, machine learning is also constrained by factors like model capacity, access to abundant labeled data, iterations required for model convergence, etc. So we hypothesize that a training regimen along similar lines may lead to an improved acquisition of meta-model. However, an unknown factor for imparting this learning strategy is the notion of ``importance."  To this end,  we propose a task attended meta-training curriculum that employs an attention module that learns to assign weights to the tasks of a batch with experience. The attention module is parametrized as a neural network that takes meta-information in terms of the model's performance  on tasks in a batch as input and learns to associate weights to each of the tasks in the batch according to their contribution in improving the meta-model.

Overall, we make the following contributions,
\begin{itemize}
    \item We design a batch-mode parametric optimizer (MetaLSTM++) that circumvents oscillations in the optimization trajectory and overcomes bias of the meta-model towards the last task. We experimentally show its merit on Omniglot, miniImageNet, and tieredImageNet datasets.
    \item We propose a task attended meta-training strategy wherein different tasks of a batch are weighted according to their ``importance" defined by the attention module. This attention module is a standalone unit that can be integrated into any batch episodic training regimen.
    \item We conduct extensive experiments on miniImageNet, and tieredImageNet datasets, and comparisons of the ML algorithms with their non-task-attended counterparts to validate the effectiveness of the task attention module.
\end{itemize}
\section{Related Work}



ML literature is profoundly diverse and may broadly be classified into \textit{metric approaches} \cite{vinyals2016matching,DBLP:conf/nips/SnellSZ17,DBLP:conf/cvpr/SungYZXTH18,DBLP:conf/nips/HouCMSC19,koch2015siamese,DBLP:journals/tip/DasL20}, \textit{initialization approaches} \cite{DBLP:conf/icml/FinnAL17,DBLP:journals/corr/abs-1803-02999,DBLP:conf/nips/FinnXL18,DBLP:conf/nips/ZhangCGBS18,DBLP:conf/iclr/AntoniouES19,DBLP:conf/cvpr/LeeMRS19,DBLP:journals/corr/LiZCL17,DBLP:conf/cvpr/JamalQ19,rajeswaran2019meta}, \textit{optimization approaches} \cite{DBLP:conf/iclr/RaviL17,andrychowicz2016learning,DBLP:conf/icml/ChenHCDLBF17,DBLP:conf/icml/WichrowskaMHCDF17} and \textit{model approaches} \cite{santoro2016meta,munkhdalai2017meta,DBLP:conf/nips/OreshkinLL18,DBLP:conf/iclr/MishraR0A18} depending on the way in which meta-knowledge is accumulated. \textit{Metric approaches} learn an embedding from input data and design kernel functions
to classify the query images by finding the maximum similarity image in the support set. \textit{Initialization approaches} learn an optimal prior on model parameters. The model is thus generalizable to new tasks drawn from the same distribution. \textit{Optimization approaches} learn parametric optimizers to traverse the loss surfaces of tasks during training and guide the model along the loss surfaces of newly sampled tasks from the same distribution. \textit{Model approaches} employ an external memory to store the meta-information gathered from the seen tasks and use it to generalize to unseen tasks. In this work, we focus on initialization and optimization approaches as they are the widely used episodic training strategies for few-shot learning. Prior work \cite{DBLP:conf/cvpr/SunLCS19,DBLP:journals/corr/abs-1910-03648,DBLP:journals/corr/abs-2007-06240} attempt to leverage the inequality/diversity of tasks in a batch. They characterize the importance of a task by virtue of its explicitly defined ``hardness" at a class level and are sensitive to hyperparameters.  Our work is different as we do not hand-craft the notion of ``importance". We, instead, define the importance at a task-level and incorporate its dependence on the meta-model's parameters. 
Our proposed approach is also comparable against a meta-training curriculum \cite{DBLP:conf/cvpr/JamalQ19}, that enforces equity across the tasks in a batch - TAML . This procedure counters adaptation, leading to slow and unstable training largely depending on the hyperparameter. We show that weighing the tasks according to their ``importance" and hence utilizing the diversity present in a batch offers better performance over enforcing equity in a batch of tasks.
\section{Background}
\subsection{Notations}
In a typical ML setting, the principal dataset $\mathcal{D}$ is divided into disjoint meta-sets $\mathcal{M}$ (meta-train set), $\mathcal{M}_v$ (meta-validation set) and $\mathcal{M}_t$ (meta-test set) for training the model, tuning its hyperparameters and evaluating its performance, respectively. Every meta-set is a collection of tasks $\Tau$ drawn from the task distribution $P(\Tau)$. Each task $\Tau_i$ consists of support $D_i =\{ \{x^c_k, y^c_k \}_{k=1}^K\}_{c=1}^N$ and query set $D_i^* =\{ \{x^{*c}_q, y^{*c}_q \}_{q=1}^Q\}_{c=1}^N$ where $(x,y)$ is a (sample, label) pair and $N$ is the number of classes, $K$ is the number of samples belonging to each class in the support set and $Q$ is the number of samples corresponding to each class in the query set. The meta-train set $\mathcal{M}$ can be written as $\{(D_i,D_i^*)\}_{i=1}^M$, where $M$ is the total number of tasks. The parameters of the meta-model are represented using $\theta$ and the base-model parameters' corresponding to each task $\Tau_i$ is $\phi_i$. 
\subsection{ Meta-knowledge as an Optimal Initialization} 
When meta-knowledge is an optimal prior on the model parameters learned through the experience over various tasks, it is enforced to be close to each individual training tasks' optimal parameters. A model initialized with such an optimal prior quickly adapts to unseen tasks from the same distribution during meta-testing. 
\textbf{MAML} \cite{DBLP:conf/icml/FinnAL17} employs a nested iterative process to learn the task-agnostic optimal prior $\theta$. In the inner iterations representing the task adaptation steps, $\theta$ is separately fine-tuned for each meta-training task $\Tau_i$ of a batch using $D_i$ to obtain $\phi_i$ through gradient descent on the train loss $L_i$. Specifically, $\phi_i$ is initialized as $\theta$ and updated using $\phi_i \leftarrow \phi_i - \alpha \nabla_{\phi_i}L_i(\phi_i)$, $T$ times resulting in the adapted model $\phi_i^T$. In the outer loop, meta-knowledge is gathered by optimizing $\theta$ over loss $L^*_i$ computed with the task adapted model parameters $\phi_i^T$ on query dataset $D^*_i$. Specifically, during meta-optimization $\theta \leftarrow \theta-\beta \nabla_{\theta} \sum_{i=1}^B L^*_i(\phi_i^T)$ using a task batch of size $B$. \textbf{MetaSGD} \cite{DBLP:journals/corr/LiZCL17} improves upon MAML by learning parameter-specific learning rates $\alpha$ in addition to the optimal initialization in a similar nested iterative procedure. Meta-knowledge is gathered by optimizing $\theta$ and $\alpha$ in the outer loop using the loss $L^*_i$ computed on query set $D^*_i$. Specifically, during meta-optimization $(\theta,\alpha) \leftarrow (\theta ,\alpha) - \beta \nabla_{(\theta , \alpha)} \sum_{i=1}^B L^*_i(\phi_i^T)$. Learning dynamic learning rates for each parameter of a model makes MetaSGD faster and more generalizable than MAML. A single adaptation step is sufficient to adjust the model towards a new task. \textbf{TAML} \cite{DBLP:conf/cvpr/JamalQ19} suggests that the optimal prior learned by MAML may still be biased towards some tasks. They propose to reduce this bias and enforce equity among the tasks by explicitly minimizing the inequality among the performances of tasks in a batch. The inequality defined using statistical measures such as Theil index, Atkinson index, Generalized entropy index, and Gini coefficient among the performances of tasks in a batch is used as a regularizer while gathering the meta-knowledge. For the baseline comparison, in our experiments, we use the Theil index for TAML owing to its average best results. Specifically during meta-optimization $\theta \leftarrow \theta-\beta \nabla_{\theta} \left[  \sum_{i=1}^B L^*_i(\phi_i^T)+ \lambda \left\{ \dfrac{ L_i^*(\theta)}{\Bar{L}^*(\theta)} \ln \dfrac{L_i^*(\theta)}{\Bar{L}^*(\theta)}\right \} \right]$  (for TAML-Theil Index) where $B$ is the number of tasks in a batch, $L^*_i$ is loss of task $\Tau_i$ on query set ${D}^*_i$ and $\Bar{L}^{*}$ is the average test loss of a batch of tasks. As TAML enforces equity of the optimal prior towards meta-train tasks, it counters the adaptation, which leads to slow and unstable training largely dependent on $\lambda$. 

\subsection{Meta-knowledge as a Parametric Optimizer} 
A regulated gradient-based optimizer gathers the task-specific and task-agnostic meta-knowledge to traverse the loss surfaces of tasks in the meta-train set during meta-training. A base model guided by such a learned parametric optimizer quickly finds the way to minima even for unseen tasks sampled from the same distribution during meta-testing. \textbf{MetaLSTM} \cite{DBLP:conf/iclr/RaviL17} is a recurrent parametric optimizer $\theta$ that mimics the gradient-based optimization of a base model $\phi$. This recurrent optimizer is an LSTM \cite{hochreiter1997long} and is inherently capable of performing two-level learning due to its architecture. During adaptation of $\phi_i$ on $D_i$,  $\theta$ takes meta information of $\phi_i$ characterized by its current loss $L_i$ and gradients $\nabla_{\phi_ {i}}(L_i)$ as input and outputs the next set of parameters for $\phi_i$. This adaptation procedure is repeated $T$ times resulting in the adapted base-model $\phi_i^T$. Internally, the cell state of $\theta$ corresponds to $\phi_i$, and the cell state update for $\theta$ resembles a learned and controlled gradient update. The emphasis on previous parameters and the current update is regulated by the learned forget and input gates respectively. While adapting $\phi_i$ to $D_i$, information about the trajectory on the loss surface across the adaptation steps is captured in the hidden states of $\theta$, representing the task-specific knowledge.  During meta-optimization, $\theta$ is updated based on the loss $L_i^*$ of model computed on query set $D^*_i$ to garner the meta-knowledge across tasks. Specifically, during meta-optimization,  $\theta \leftarrow \theta-\beta \nabla_{\theta} L^*_i(\phi_i^T)$.\\
 \begin{figure}[t]
\centering
 \includegraphics[width =0.5\linewidth]{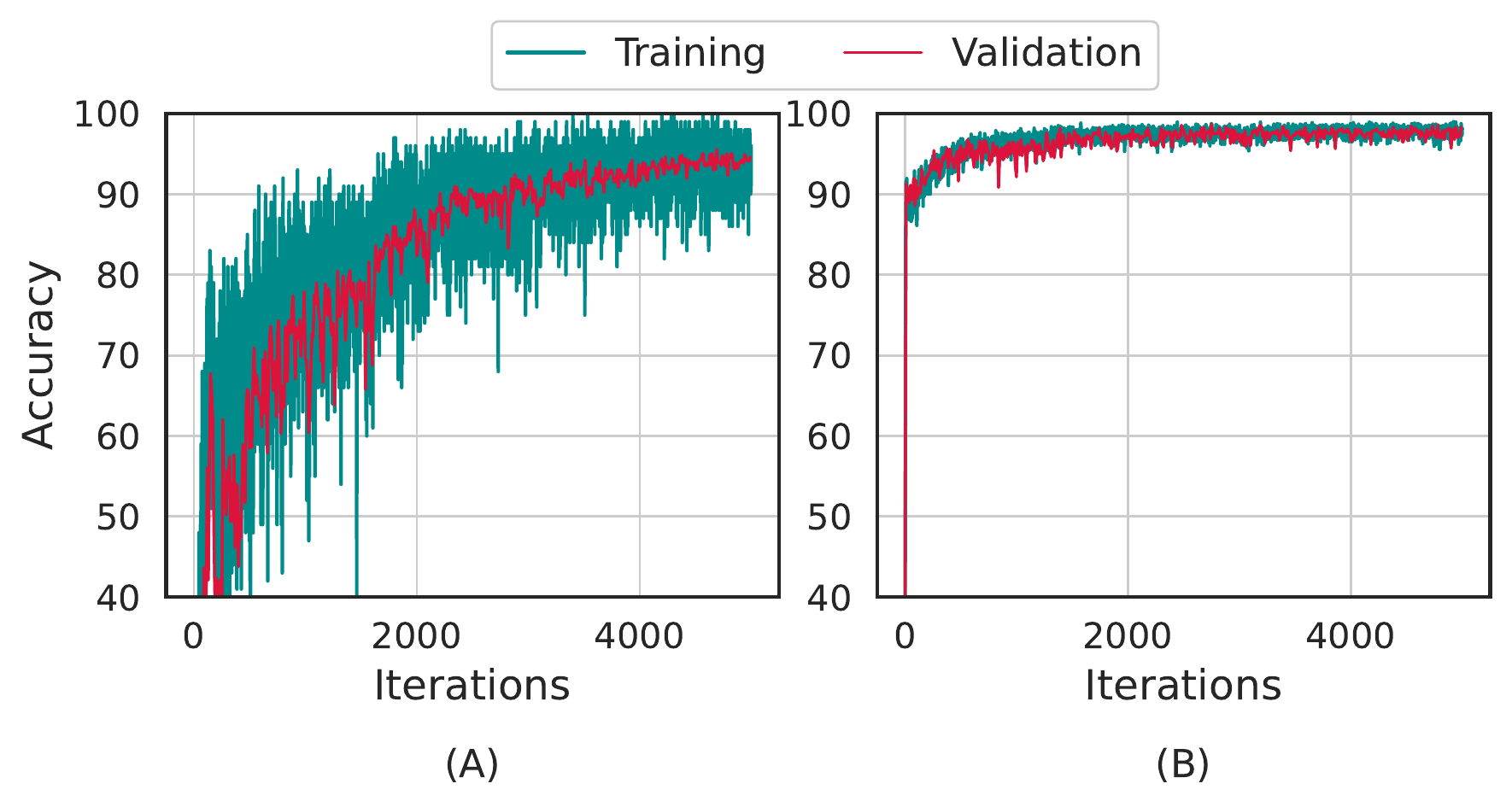}
  \caption{Oscillations in MetaLSTM (A) compared to MetaLSTM++ (B).}
\label{fig:oscillations}
\end{figure}
\textbf{ MetaLSTM++} A caveat of MetaLSTM is the sequential update to the parametric optimizer $\theta$ after adapting to each task. As a result, the parametric optimizer traverses the loss surface in an ordered sequence of task-specific optima. This leads to a longer and oscillatory optimization trajectory as shown in Figure \ref{fig:oscillations} and bias of $\theta$ towards the final task. We propose to overcome this bottleneck by learning $\theta$ according to the training procedure of optimal initialization ML approaches, which we term as MetaLSTM++. Unlike MetaLSTM, the meta-knowledge $\theta$ of MetaLSTM++ is updated based on the average test loss of the tasks in a batch. This is intuitive as a batch of tasks may better approximate the task distribution than a single task. The batch update on $\theta$ makes the optimization trajectory smooth, short, and robust to task order (Figure \ref{fig:oscillations}). Specifically, during meta-optimization $\theta \leftarrow \theta-\beta \nabla_{\theta} \sum_{i=1}^B L^*_i(\phi_i^T)$.

\section{Task Attention in Meta-learning}
A common assumption under the batch wise episodic training regimen adopted by ML is that each task in a batch has an equal contribution in improving the learned meta-knowledge. However, this need not always be true. It is likely that given the current configuration of the meta model, some tasks may be more important for the meta-model's learning. A contributing factor to this difference is that tasks sampled from complex data distributions can be profoundly diverse. The diversity and latent properties of the tasks coupled with the model configuration may induce some tasks to be better aligned with the optimal meta-knowledge than others. The challenging aspect in the meta-learning setting is to define the ``importance"  and associate weights to the tasks of a batch in proportion to their contribution to improving the meta-knowledge. As human beings, we \textit{learn} to  associate importance to events subjective to meta-information about the events and prior experience. This motivates us to define a learnable module that can learn to map the meta-information of tasks to their importance weights. Specifically, we learn a task attention module parameterized by $\delta$, which attends to the tasks that contribute more to the model's learning.

Given a task-batch $\{\Tau_i\}_{i=1}^B$, the task attention module takes as input meta-information about each task $(\Tau_i)$ in the batch, defined as the four tuple below:
\begin{equation}
    \mathcal{I} = \left\{ \ \left( \  ||\nabla_{\phi_i} L_i^*(\phi_i^T)||, L_i^{*T}, A_i^{*T}, \dfrac{L_i^{*T}}{L_i^{*0}} \ \right) \ \right\}_{i=1}^B
\end{equation}
where $||\nabla_{\phi_i} L_i^*(\phi_i^T)||$, $L_i^{*T}$ and $A_i^{*T}$ denote the norm of gradient, test loss, and the accuracy of the adapted model parameters of the $i^{th}$ task in the batch respectively. The ratio of the test loss of the model post and prior adaptation, $\dfrac{L_i^{*T}}{L_i^{*0}}$ helps capture the relative progress achieved on each task by the meta-model $\theta$. 

The objective of the task attention module is to learn the relative importance of each task in the batch for the meta-model's learning. Thus the output of the module is a $B-$dimensional vector $\textbf{w}=[w_1,\ldots,w_B]$, ($\sum_{i=1}^B w_i = 1$) quantifying the attention-score (weight - $w_i$) for each task. The attention vector $\textbf{w}$ is multiplied with the corresponding task losses of the adapted models $L_i^*{ (\phi_i^T)}$ on the held-out datasets $D_i^*$ to update the meta-model parameter $\theta$: 
 \begin{equation}
    \theta^{t+1} \leftarrow \theta^t - \beta \nabla_{\theta^t} \sum_{i=1}^B w_i L^*_i (\phi_i^T)
\end{equation}

After the meta-model is updated using the weighted task losses, we evaluate the goodness of the generated attention weights. We sample a new batch of tasks $\{D_j,D_j^*\}_{j=1}^B$ and adapt a base-model $\phi_j$ using the updated meta-model $\theta^{t+1}$ on the train data $\{D_j\}$ of each task. 
The mean test loss of the adapted models $\{\phi_j^T\}_{j=1}^B$ reflect the goodness of the weights assigned by the attention-module in the previous iteration. The attention module $\delta$ is thus updated using the gradients flowing back into it w.r.t to this test-loss. The attention network is trained simultaneously with the  meta-model in end to end fashion using the update rule: 

\begin{equation}
    \delta^{t+1} \leftarrow \delta^t - \gamma \nabla_{\delta^t} \sum_{j=1}^B L^*_j ({\phi_j^T}),
    \text{ where } {\phi_j^T} \text{ is adapted from } \theta^{t+1}
\end{equation}
\begin{figure}[h]
    \centering
    \includegraphics[width = 0.8\linewidth]{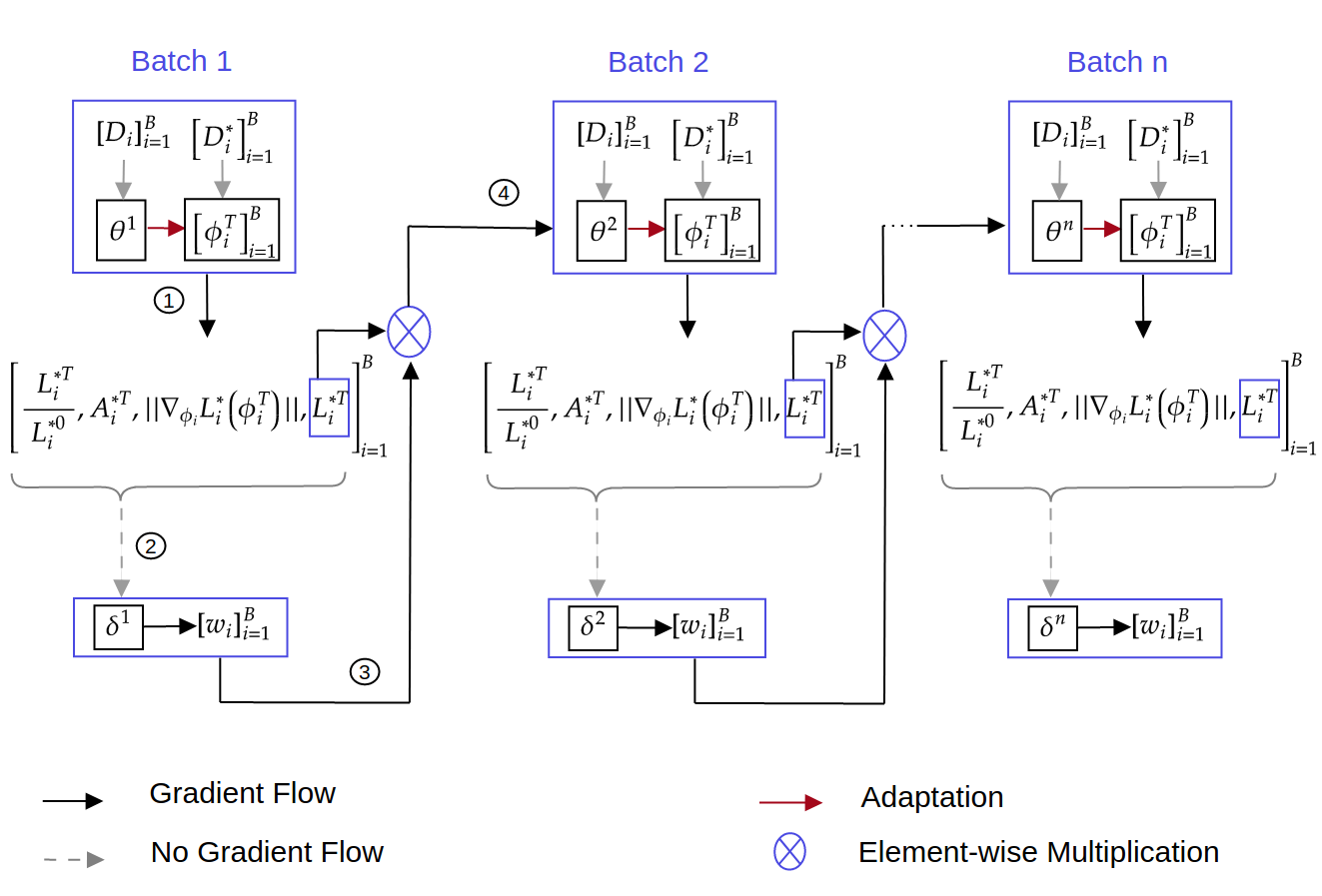}
    \caption{Computational Graph of the forward pass of the meta-model using task attended meta-training curriculum. The output of this procedure is a meta-model $\theta^n$. Gradients are propagated through solid lines and restricted through dashed lines.}
    \label{fig:gradient_flow}
\end{figure}
\subsection{Task Attended Meta-Training Algorithm}
\begin{algorithm}[h]
\SetAlgoLined
\KwIn{Dataset: $\mathcal{M}=\{D_i,D_i^*\}_{i=1}^M$}
\myinput{Models : Meta-model $\theta$, Base-model $\phi$, Attention-module $\delta$}
\myinput{Parameters : Iterations $n_{iter}$, Batch-size $B$, Adaptation-steps $T$} 
\myinput{Learning-rates : $\alpha$, $\beta$, $\gamma$}
\KwOut{Meta-model : $\theta$}
\textbf{Initialization:} $\theta, \delta \leftarrow $ \text{Random Initialization} \\
\For{iteration in $n_{iter}$}{ 
$\{\Tau_i\}_{i=1}^B= \{D_i,D_i^*\}_{i=1}^B \leftarrow$ Sample-task-batch($\mathcal{M}$) \\
\For{all $\Tau_i$}{
$\phi_i^0 \leftarrow \theta$ \\
$L_i^{*0},\_ \leftarrow evaluate (\phi_i^0,D_i^* )$ \\
$\phi_i^T=adapt(\phi_i^0, D_i)$ \\
$L_i^{*T},A_i^{*T}   \leftarrow evaluate (\phi_i^T,D_i^* )$ 
}  
$\left[w_i\right]_{i=1}^B \leftarrow Attention\_module \left(\left[\dfrac{L_i^{*T}}{L_i^{*0}},A_i^{*T}, ||\nabla_{\phi_i} L_i^*(\phi_i^T)|| , L_i^{*T} \right]_{i=1}^B \right)$ \\ 
$\theta \leftarrow \theta - \beta \nabla_\theta \sum_{i=1}^B w_i L_i^* (\phi_i^T)$ \\
$\{D_j,D_j^*\}_{j=1}^B \leftarrow$ Sample-task-batch($\mathcal{M}$) \\
\For{all $\Tau_j$}{
$\phi_j^0 \leftarrow \theta$ \\
$\phi_j^T=adapt(\phi_j^0, D_j)$ 
}
$\delta \leftarrow \delta - \gamma \nabla_\delta \sum_{j=1}^B L_j^* (\phi_j^T)$
}
\textbf{Return} $\theta $ 
\caption{Task Attended Meta-Training}
\label{alg:Task Attended Metalearning}
\end{algorithm}

We demonstrate a meta-training curriculum using the proposed task attention in Figure \ref{fig:gradient_flow}, formally summarized in Algorithm \ref{alg:Task Attended Metalearning}. As with the classical meta-training process, we first sample a batch of tasks from the task distribution. For each task $\Tau_i$, we adapt the base-model $\phi_i$ using the train data $\{D_i\}_{i=1}^B$ for $T$ time-steps (line 10). Meta-information about the adapted models for each task is then computed, comprising of the loss $L_i^{*T}$, the accuracy $A_i^{*T}$, the loss-ratio $\dfrac{L_i^{*T}}{L_i^{*0}}$ and gradient norm on test data $\{D_i^*\}_{i=1}^B$. The meta-information corresponding to each task in a batch is given as input to the task attention module (Figure \ref{fig:gradient_flow} - Label: \circled{2}) which outputs the attention vector (line 13). The attention vector in combination with test losses $\{L^*_i\}_{i=1}^B$ is used to update meta-model parameters $\theta$  (line 14, Figure \ref{fig:gradient_flow} - Label: \circled{4}). We sample a new batch of tasks $\{D_j, D_j^*\}_{j=1}^B$ and adapt the base-models $\{\phi_j^T\}_{j=1}^B$ using the updated meta-model. We compute the mean test loss over the adapted base-models $\{L_j^* (\phi_j^T)\}_{j=1}^B$, which is then used to update the parameters of the task attention module $\delta$ (lines 15-20).


The attention network is designed as a stand-alone module to learn the mapping from the meta-information space to the importance of tasks in a batch. Thus, it is important to decouple the learning of the attention network from that of the meta-model. The parameters of the meta-model $\theta$ should not be directly dependent on that of the task attention module $\delta$. If information about the attention network's learning flow into the meta-model, it is possible that the meta-model's learning shifts towards fooling the task attention network to output weights that are biased. We prevent this occurrence by enforcing $\nabla_{\theta}  w_i L^*_i (\phi_i^T) = w_i\nabla_{\theta} L^*_i (\phi_i^T)$. We restrict the flow of gradients to the meta-model through task attention module. Figure \ref{fig:gradient_flow} demonstrates the paths along which gradient backflow is restricted and permitted as dashed and solid lines respectively.

\section{Experiments and Results}

We consider different few-shot learning settings on the benchmark datasets - Omniglot, miniImageNet, and tieredImageNet to test the effectiveness of the proposed approach. All the experimental results and comparisons correspond to our re-implementation of the methods. We perform individual hyperparameter tuning for all the models, over the same hyperparameter space to ensure a fair comparison. The architecture and hyper-parameter details for each are provided in the implementation details. The source code is publicly available \footnote{\url{https://github.com/selective-task-attention/TaskAttention}}

\subsection{Datasets and Implementation Details}
\textbf{Omniglot} \cite{lake2015human} dataset comprises 1623 characters from 50 different alphabets, where each character is written by 20 different individuals. The images are downsampled to 28 $\times$ 28. The standard split consists of 1200 classes for meta-training and the rest for meta-testing \cite{DBLP:conf/icml/FinnAL17}. We follow the standard split but keep 220 classes from the meta-training split to tune the models' hyperparameters. \textbf{miniImageNet} \cite{vinyals2016matching} comprises 600 color images of size 84 $\times$ 84 from each of 100 classes sampled from the Imagenet dataset. The 100 classes are split into 64, 16  and 20 classes for meta-training, meta-validation and meta-testing respectively. \textbf{tieredImageNet} \cite{DBLP:conf/iclr/RenTRSSTLZ18} is a more challenging benchmark for few-shot image classification. It contains 779,165 color images sampled from 608 classes of Imagenet and are grouped into 34 super-classes. Each super-classes is divided into 20, 6, and 8 disjoint sets for meta-training, meta-validation, and meta-testing. In line with the state of the art literature \cite{DBLP:journals/corr/abs-1910-03648}, we also only use miniImageNet and tieredImageNet for evaluating the effectiveness of the proposed attention module as they are more challenging datasets comprising of highly diverse tasks.

We use the architecture from \cite{DBLP:conf/icml/FinnAL17} for the base model and a two-layer LSTM \cite{DBLP:conf/iclr/RaviL17} for the parametric optimizer. The task attention module is a ReLU activated neural network with a 1 $\times$ 1 convolutional layer followed by 2 fully connected layers with 32 neurons, and finally a softmax activation to generate the attention weights. We perform a grid search over 30 different configurations for 5000 iterations to find the optimal hyper-parameters for each setting. The search space is shared across all meta-training algorithms. The meta, base and attention model learning rates are sampled from a log uniform distribution in the ranges $\left[1e^{-4} , 1e-2 \right]$, $\left[1e^{-2} , 5e^{-1} \right]$ and $\left[1e^{-4} , 1e^{-2} \right]$ respectively. The hyperparameter $\lambda$ for TAML(Theil) is sampled from a log uniform distribution over the range of $\left[1e^{-2} , 1 \right]$. The number of adaptation steps is fixed to 5 for all settings except for 10-way 5-shot setting, where we use 2 adaptation steps owing to the computational expenses. The meta-batch size is set to 4 for miniImageNet and tieredImageNet and 16 in Omniglot. All models were trained for 55000 iterations using the optimal set of hyper-parameters using an Adam optimizer\cite{DBLP:journals/corr/KingmaB14}.
\begin{table}[h]
\centering
\caption{Comparison of few-shot classification performance of MetaLSTM and MetaLSTM++ on the miniImageNet, tieredImageNet and Omniglot datasets.  We employ 5 and 10 way (1 and 5 shot) settings for miniImageNet and tieredImageNet and 20 way (1 and 5 shot) settings for Omniglot. The $\pm$ represents the 95\% confidence interval across 300 tasks. All the algorithms are rerun (denoted by *) on their optimal hyper-parameters for a fair comparison. MetaLSTM++ outperforms MetaLSTM across all settings in all datasets.}
\begin{tabular}{@{}lcccccccccccccccc@{}}
\toprule 
 & \multicolumn{4}{c}{ \textbf{Test Accuracy (\%)}} \\ 
 \cmidrule(lr){2-5} 

  \multicolumn{1}{l}{ } & \multicolumn{2}{c}{5-Way} & \multicolumn{2}{c}{10-Way} \\ 
  \cmidrule(lr){2-3} 
  \cmidrule(lr) {4-5}
  \multicolumn{1}{l}{ \textbf{Model}} & \multicolumn{1}{c}{1 Shot} & \multicolumn{1}{c}{5 Shot } & \multicolumn{1}{c}{1 Shot} & \multicolumn{1}{c}{5 Shot } \\ 
\midrule
  & {  }  & \textbf{ miniImageNet}  & {  }  & {  } \\
\midrule
MetaLSTM$^*$                     & 41.48 $\pm$ 1.02 & 58.87 $\pm$ 0.94  & 28.62 $\pm$ 0.64 & 44.03 $\pm$ 0.69  \\ 
\addlinespace
\textbf{MetaLSTM++ }         & \textbf{48.00 $\pm$ 0.19} & \textbf{62.73 $\pm$ 0.17 }  & \textbf{ 31.16 $\pm$ 0.09} & \textbf{45.46 $\pm$ 0.10}  \\ 
\midrule
  & {  }  & \textbf{ tieredImageNet}  & {  }  & {  } \\
\midrule
\addlinespace
MetaLSTM$^*$                     & 37.00 $\pm$ 0.44 & 59.83  $\pm$ 0.25  &  29.80 $\pm$ 0.28 & 39.28 $\pm$ 0.13   \\ 
\addlinespace
\textbf{MetaLSTM++   }       & \textbf{47.60 $\pm$ 0.49} &\textbf{ 63.24 $\pm$ 0.25} &  \textbf{30.70 $\pm$ 0.27} & \textbf{47.97 $\pm$ 0.16 } \\ 
\addlinespace
\midrule
  & {  }  & \textbf{ Omniglot}  & {  }  & {  } \\
\midrule
\multicolumn{1}{l}{ } & \multicolumn{2}{c}{20-Way 1-Shot} & \multicolumn{2}{c}{20-Way 5-Shot} \\
\cmidrule(lr){2-3} 
\cmidrule(lr) {4-5}
MetaLSTM$^*$    & \multicolumn{2}{c}{90.63 $\pm$ 0.83} &  \multicolumn{2}{c}{97.11 $\pm$ 0.24}   \\ 
\addlinespace
\textbf{MetaLSTM++   }        & \multicolumn{2}{c}{\textbf{96.50 $\pm$ 0.42}} & \multicolumn{2}{c}{\textbf{98.41 $\pm$ 0.31}} \\ 
\addlinespace
\bottomrule
\end{tabular}
\\
\label{MetaLSTM++MI+TI}
\end{table}

\subsection{Comparison of MetaLSTM++ with MetaLSTM}
We experiment on the more challenging 20-way (1 and 5 shot) setting for the Omniglot dataset and 5 and 10 way (1 and 5 shot) setting for miniImageNet and tieredImageNet datasets to study the importance of batch updates in the parametric optimizer. The mean meta-test accuracies for MetaLSTM and MetaLSTM++ for each of the above setting are summarized in Table \ref{MetaLSTM++MI+TI}. We can observe from the results that on all the three datasets MetaLSTM++ outperforms MetaLSTM in both the 1-shot and 5-shot settings. The result validates the effectiveness of batch-wise meta-training for the parametric optimizer.
\begin{table}[h!]
\centering
\caption{Comparison of few-shot classification performance of vanilla ML algorithms with their task attended versions on the miniImageNet and tieredImageNet dataset for 5 and 10 way (1 and 5 shot) settings. The $\pm$ represents the 95\% confidence intervals over 300 tasks. All the algorithms are rerun (denoted by *) on their optimal hyper-parameters for a fair comparison. Attention-based ML algorithms perform better than their corresponding vanilla approaches across all the settings. Further, TA-MAML performs better than TAML across all settings and datasets.}
\begin{tabular}{@{}lcccccccccccccccc@{}}
\toprule 
 & \multicolumn{4}{c}{ \textbf{Test Accuracy (\%)}} \\ 
 \cmidrule(lr){2-5}

  \multicolumn{1}{l}{ } & \multicolumn{2}{c}{5-Way} & \multicolumn{2}{c}{10-Way} \\ 
  \cmidrule(lr){2-3} 
  \cmidrule(lr) {4-5}

  \multicolumn{1}{l}{ \textbf{Model}} & \multicolumn{1}{c}{1 Shot} & \multicolumn{1}{c}{5 Shot } & \multicolumn{1}{c}{1 Shot} & \multicolumn{1}{c}{5 Shot } \\ 
\midrule

  & {  } &  \textbf{miniImageNet}  & {  }  & {  } \\
\midrule
MAML$^*$                     & 46.10 $\pm$ 0.19 & 60.16 $\pm$ 0.17  & 29.42 $\pm$ 0.11 & 41.98 $\pm$ 0.10  \\ 
TAML$^*$                    & 46.26 $\pm$ 0.21 & 53.40 $\pm$ 0.14   & 29.76 $\pm$ 0.11 & 36.88 $\pm$ 0.10 \\ 
\textbf{TA-MAML}   & \textbf{48.36 $\pm$ 0.23} & \textbf{ 62.48 $\pm$ 0.18
 }  & \textbf{31.15$\pm$ 0.11 } & \textbf{ 43.70 $\pm$ 0.09 } \\ \addlinespace
\midrule

MetaSGD$^*$                     & 47.65$\pm$ 0.21 & 61.60 $\pm$ 01.71   &  30.09$\pm$ 0.10  & 42.22 $\pm$ 0.11   \\ 
\textbf{TA-MetaSGD}   & \textbf{ 49.28 $\pm$ 0.20} & \textbf{63.37 $\pm$  0.16}  & \textbf{31.50$\pm$ 0.11}  & \textbf{44.06 $\pm$ 0.10 } \\ \addlinespace
\midrule

MetaLSTM++                     & 48.00 $\pm$ 0.19 & 62.73 $\pm$ 0.17   &  31.16 $\pm$ 0.09 & 45.46 $\pm$ 0.10  \\ 
\textbf{TA-MetaLSTM++}   & \textbf{ 49.18 $\pm$ 0.17} & \textbf{64.89 $\pm$  0.16 }  & \textbf{32.07$\pm$ 0.11}  & \textbf{46.66 $\pm$ 0.09} \\ \addlinespace
\midrule
  & {  }  & \textbf{ tieredImageNet}  & {  }  & {  } \\
\midrule
MAML$^*$                     & 44.40 $\pm$ 0.49 & 57.07 $\pm$ 0.22  &  27.40 $\pm$ 0.25 & 34.30  $\pm$ 0.14  \\ 
TAML$^*$                   & 46.40 $\pm$ 0.40 & 56.80  $\pm$ 0.23  &  26.40 $\pm$ 0.25 & 34.40 $\pm$ 0.15  \\ 
\textbf{TA-MAML}   & \textbf{48.40 $\pm$ 0.46} & \textbf{60.40 $\pm$ 0.25 }  & \textbf{31.00$\pm$ 0.26} & \textbf{37.60$\pm$ 0.15} \\ \addlinespace
\midrule

MetaSGD$^*$                    & 52.80 $\pm$ 0.44 & 62.35  $\pm$ 0.26  &  31.90 $\pm$ 0.27 & 44.16 $\pm$ 0.15   \\ 
\textbf{TA-MetaSGD}   & \textbf{ 56.20  $\pm$ 0.45} & \textbf{64.56 $\pm$ 0.24}  & \textbf{33.20$\pm$ 0.29}  & \textbf{47.12 $\pm$ 0.16 } \\ \addlinespace
\midrule

MetaLSTM++                     & 47.60 $\pm$ 0.49 & 63.24 $\pm$ 0.25 &  30.70 $\pm$ 0.27 & 47.97 $\pm$ 0.16  \\ 
\textbf{TA-MetaLSTM++}   & \textbf{ 49.00 $\pm$ 0.44} & \textbf{66.15 $\pm$  0.23}  & \textbf{32.10$\pm$ 0.27}  & \textbf{51.35 $\pm$ 0.17 } \\ \addlinespace
\bottomrule

\end{tabular}
\label{accuracy_table}
\end{table}

\subsection{Influence of Task Attention on Meta-Training}
As the task-attention (TA) is a standalone module, it can be integrated with any batch episodic training regimen. Thus, we investigate the performance of the models trained with the TA meta-training regimen with their non-TA counterparts. Specifically, we compare MAML, MetaSGD, and MetaLSTM++ with TA-MAML and TA-MetaSGD and TA-MetaLSTM++ respectively over 5 and 10 way (1 and 5 shot) settings on miniImageNet and tieredImageNet datasets and report the results in Table \ref{accuracy_table}. We observe that models trained with TA regimen generalize better to the unseen meta-test tasks than their non-task-attended versions across all the settings in both datasets.

We also compare the performance of TA-MAML against TAML - a meta-training regimen
that forces the meta-model to be equally close to all the tasks. 
The results, as presented in Table \ref{accuracy_table}, suggest that TA-MAML performs better than TAML on both benchmarks across all settings. Note that both TAML and TA-MAML are approaches that build upon MAML to address the inequality/diversity of tasks in a batch. Our aim is thus to compare TAML and TA-MAML and not to assess the efficacy of TAML when meta-trained by task attention procedure.

\begin{figure}[h]
\centering
\begin{tabular}{ccc}
\subfloat[MAML]{\includegraphics[width = 0.333\linewidth]{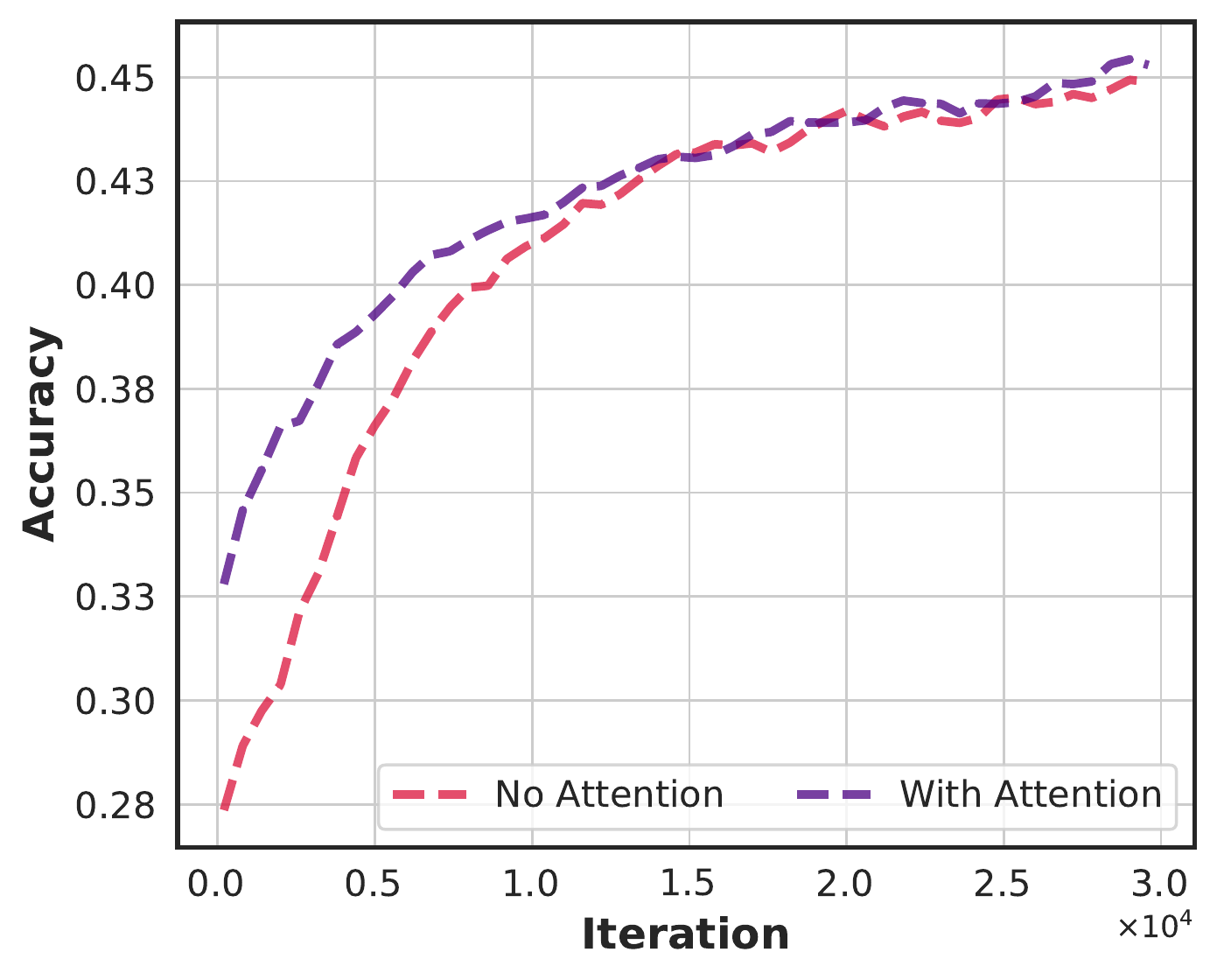}}&
\subfloat[MetaSGD ]{\includegraphics[width = 0.333\linewidth]{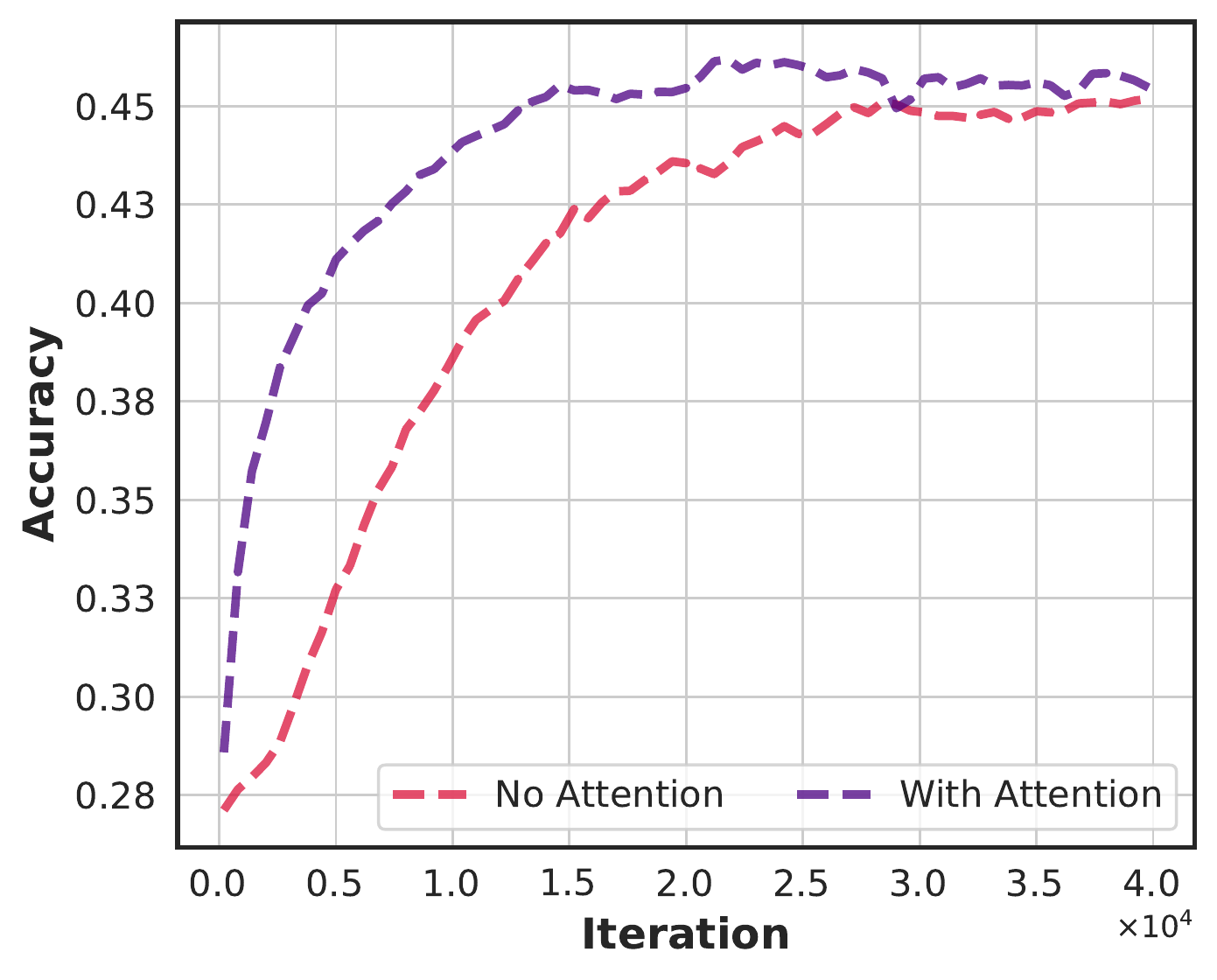}}&
\subfloat[MetaLSTM++ ]{\includegraphics[width = 0.333\linewidth]{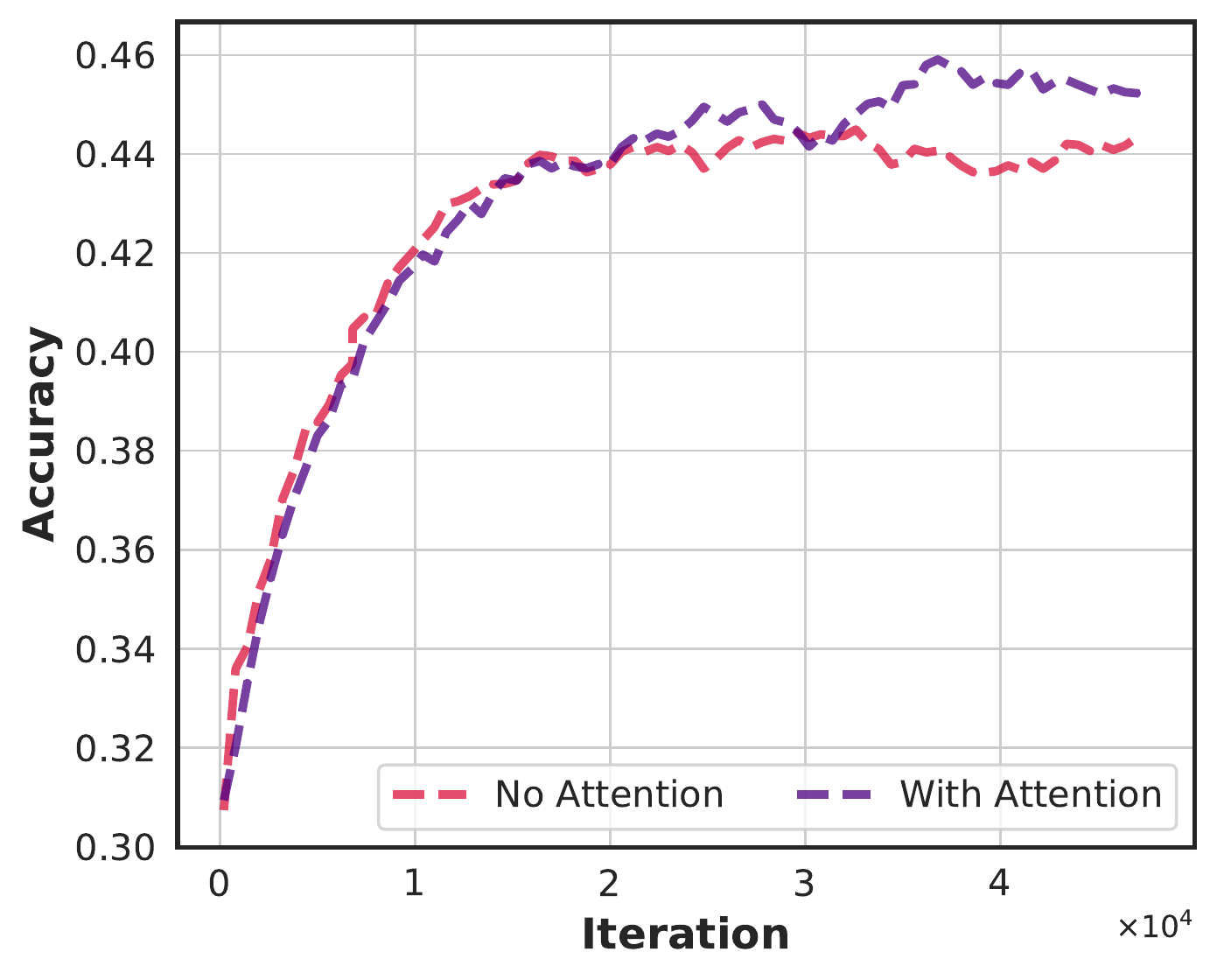}} \\
\subfloat[MAML ]{\includegraphics[width = 0.333\linewidth]{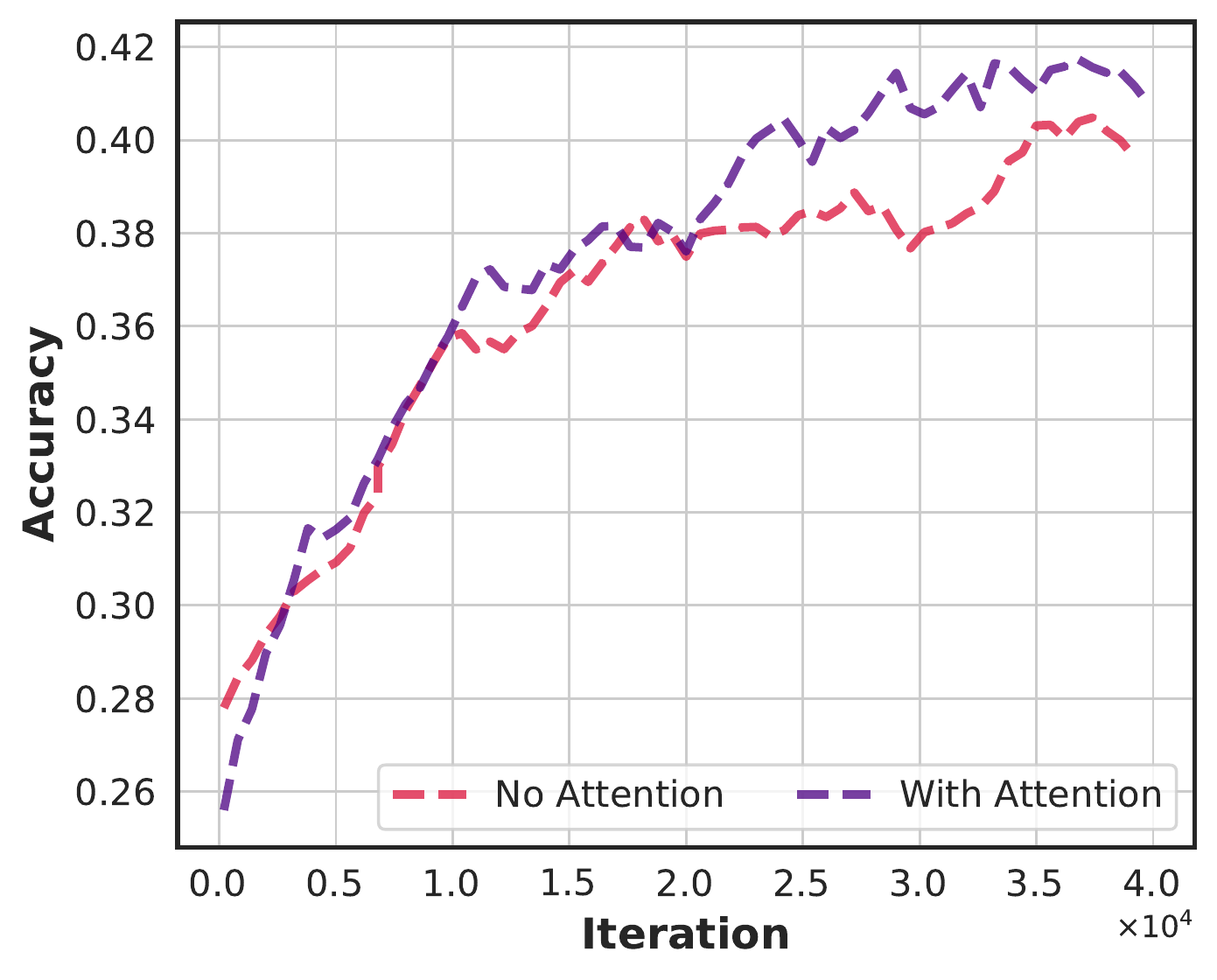}}&
\subfloat[MetaSGD ]{\includegraphics[width = 0.333\linewidth]{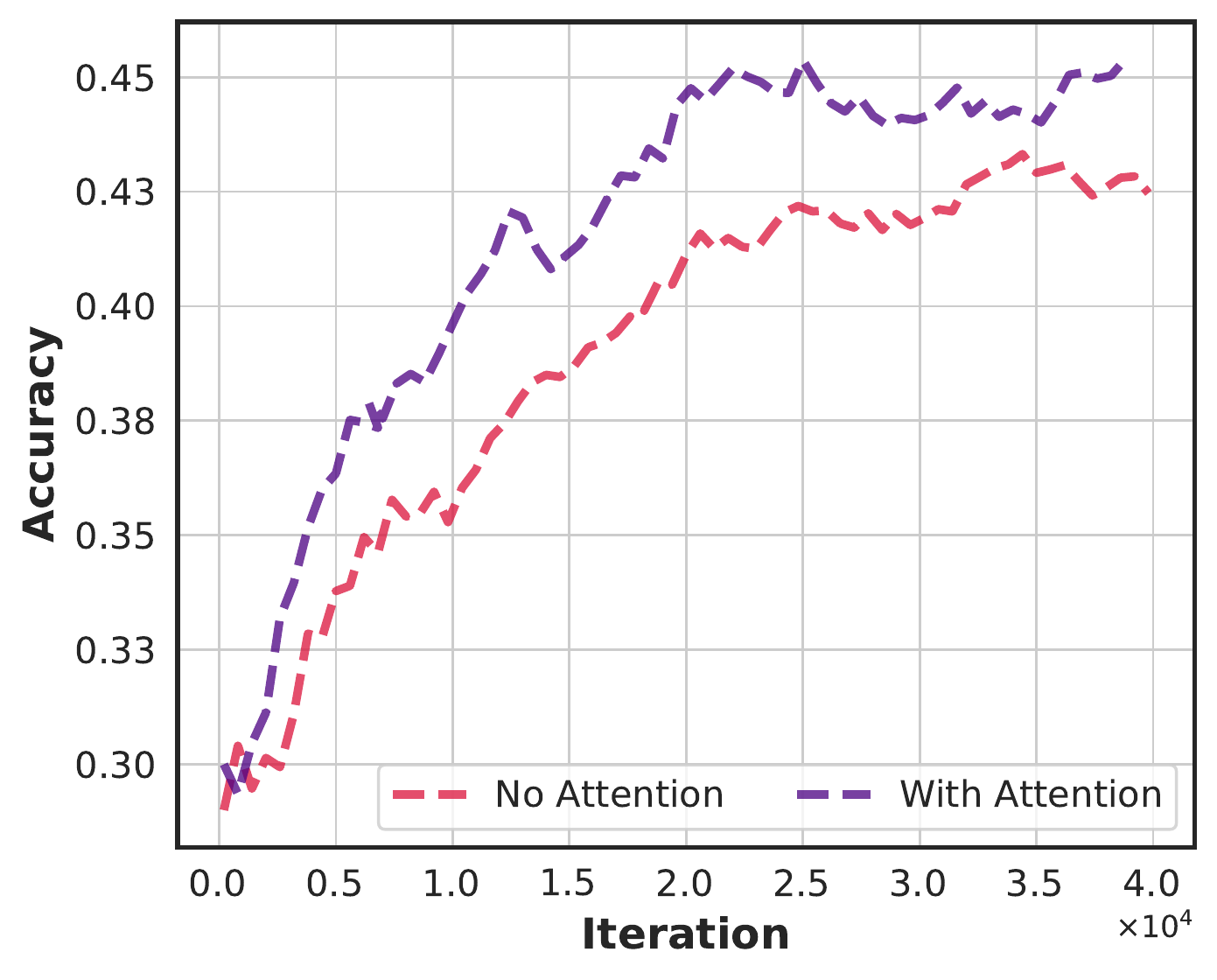}}&
\subfloat[MetaLSTM++]{\includegraphics[width = 0.333\linewidth]{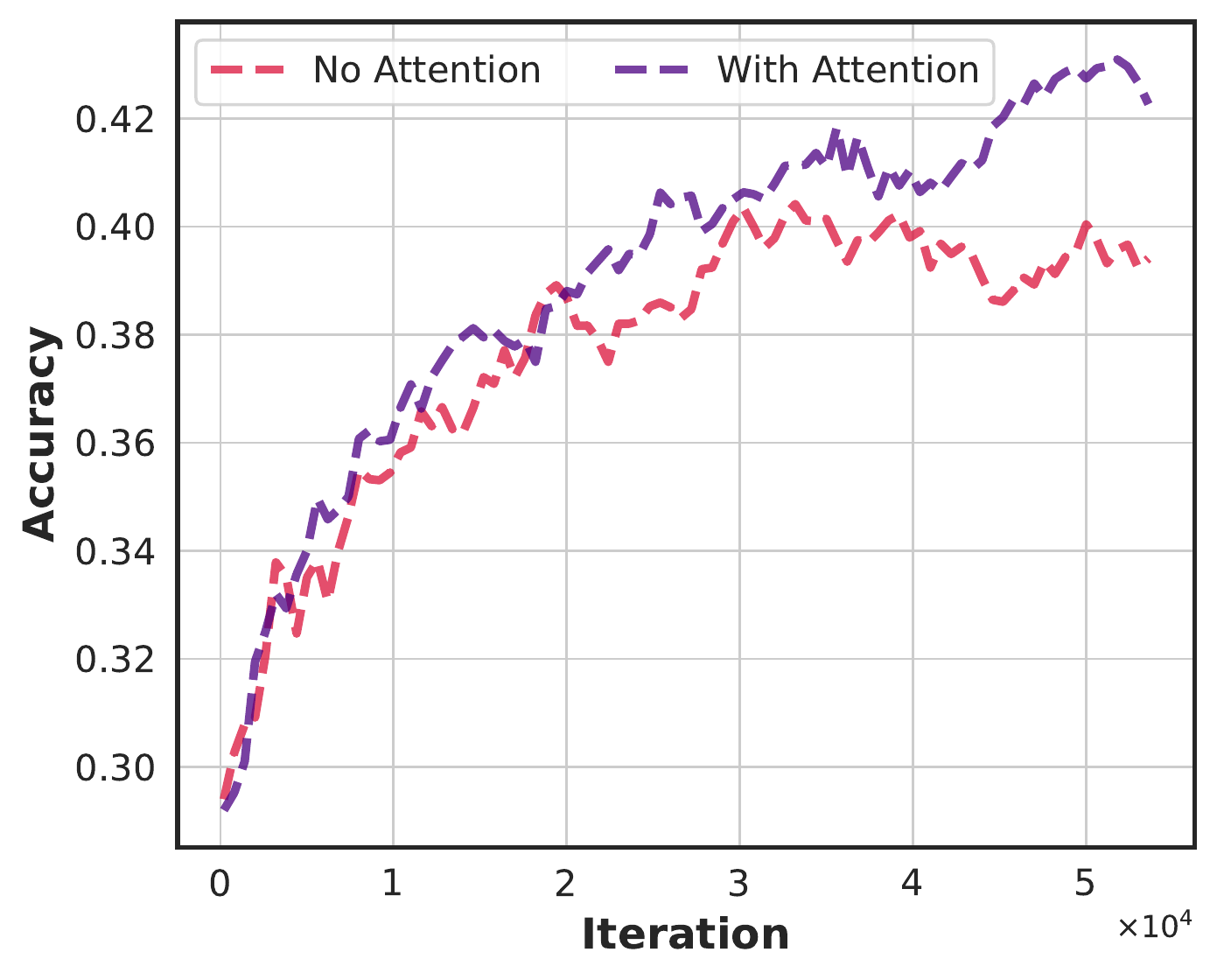}}
\end{tabular}
\caption{[Best viewed in color]  Mean validation accuracies of ML algorithms across 300 tasks with/without attention on 5-way 1-shot setting on miniImageNet and tieredImageNet. Rows represent datasets; Row-1 corresponds to miniImageNet, and Row-2 corresponds to tieredImageNet. Coloumns represent training algorithms; Coloumn-1 corresponds to MAML, Coloumn-2 to MetaSGD and Coloumn-3 to MetaLSTM++. }
\label{fig:validation_MI_TI}
\end{figure}

We also investigate the influence of the TA meta-training regimen on the model's convergence by analyzing the trend of the model's validation accuracy over iterations. Figure \ref{fig:validation_MI_TI} depicts the mean validation accuracy over 300 tasks on miniImageNet and tieredImageNet datasets for 5-way 1-shot setting across training iterations. We observe that the models meta-trained with TA regimen tend to achieve higher/at-par performance in fewer iterations than the corresponding models meta-trained with the non-TA regimen.

\begin{figure}[h]
\centering
\begin{tabular}{cc}
\subfloat[TA-MAML ]{ \includegraphics[width = 0.33\linewidth]{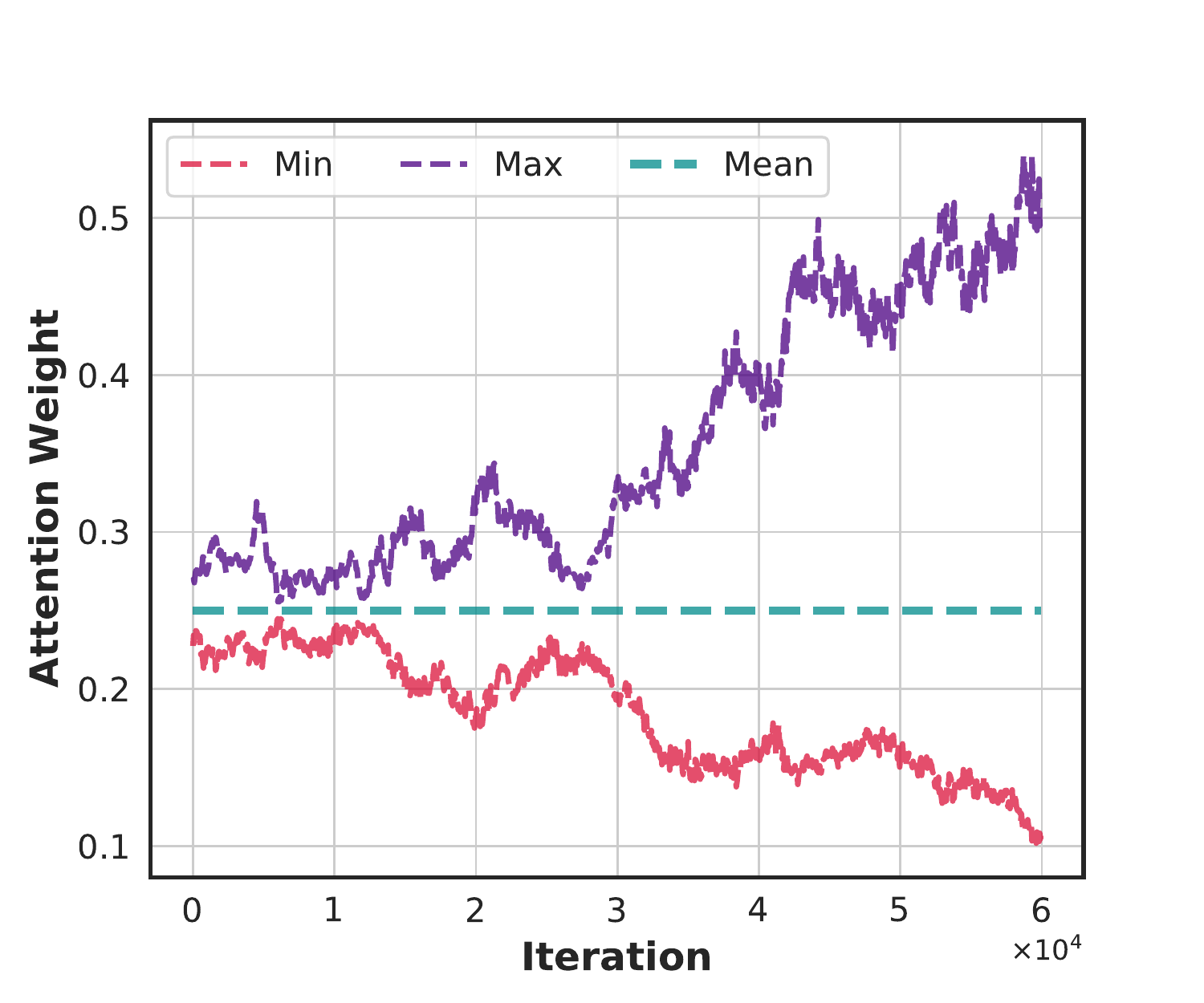}} &
\subfloat[TA-MetaLSTM++]{\includegraphics[width = 0.33\linewidth]{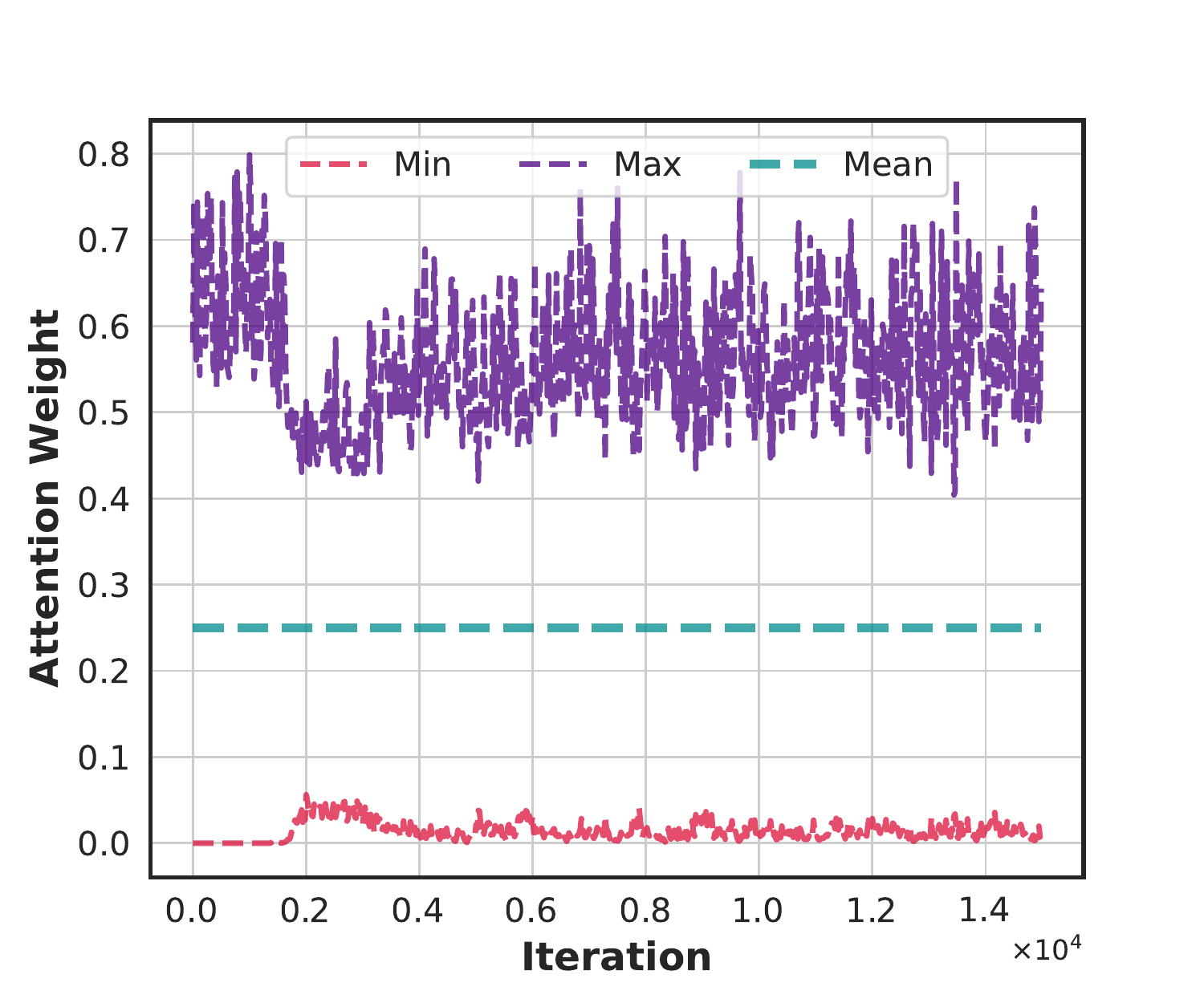}} 
\end{tabular}
 \caption{Trend of an attention vector in a 5-way 1-shot setting on miniImageNet dataset for (a) TA-MAML (b) TA-MetaLSTM++ during meta-training.}
\label{fig:attention_trend_5.1}
\end{figure}

\subsection{Analysis of Attention Network}
We investigate whether the TA module is trivially following any meta-information component for assigning weights or using some learned latent information about tasks to assign the weights. So, we first examine the trend of the attention-vector (Figure \ref{fig:attention_trend_5.1}) by plotting the maximum and the minimum attention score assigned to the tasks of a batch across iterations. We perform this study during meta-training TA-MAML and TA-MetaLSTM++ in a 5-way 1-shot setting on miniImageNet dataset.  Note that the mean attention-score is always 0.25 as we follow a meta-batch size of 4. We observe that the TA module's output follows an interesting trend in the case of TA-MAML. Initially, the TA module assigns almost uniform weights to all the tasks of a batch; however, as the iterations increase, the TA module assigns unequal scores to the tasks in a batch, preferring some over the other. This suggests that for initialization based approaches, during the initial phases of the meta-model's training all tasks have equal contribution towards learning a \textit{generic structure} of the meta-knowledge. As the meta-model's learning proceeds, learning the further \textit{fine-grained meta-knowledge structure} requires prioritizing some tasks in a batch over the others, which are potentially better aligned with learning the optimal meta-knowledge.
The trend of the attention scores in the case of TA-MetaLSTM++ (Figure \ref{fig:attention_trend_5.1}(b)) is different than the one observed in TA-MAML. We observe a high variance in the attention weights assigned, which decreases initially but then increases. While this suggests that the weights assigned are dynamic to the learning of the parametric optimizer, it also helps draw an insight that parametric optimizers can possibly learn varying knowledge about traversing loss surfaces from the individual tasks, each adding up to the optimal meta-knowledge.\\

\begin{figure}[h]
\centering
\begin{tabular}{cc}
\subfloat[Correlation Analysis]{ \includegraphics[width = 0.33\linewidth]{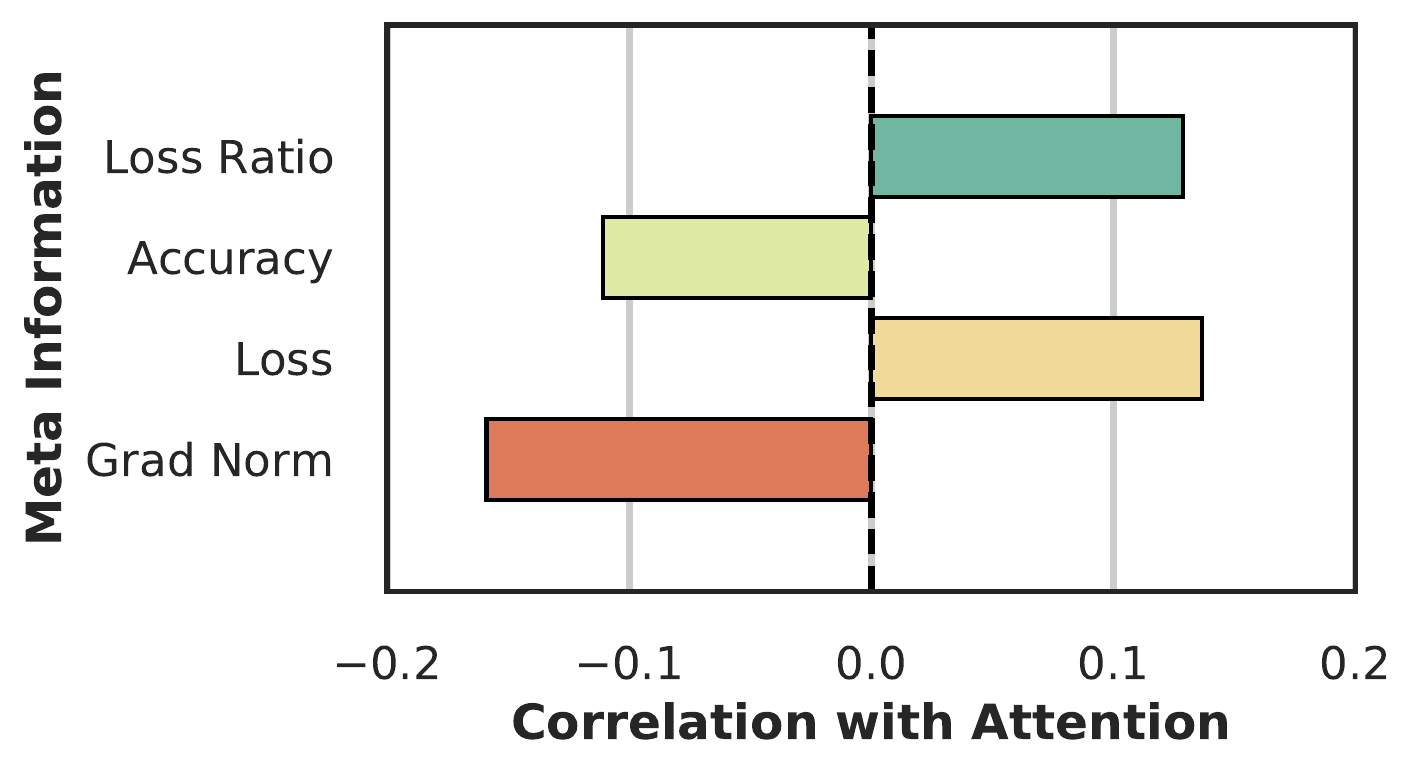}} &
\subfloat[Rank Analysis]{\includegraphics[width = 0.72\linewidth]{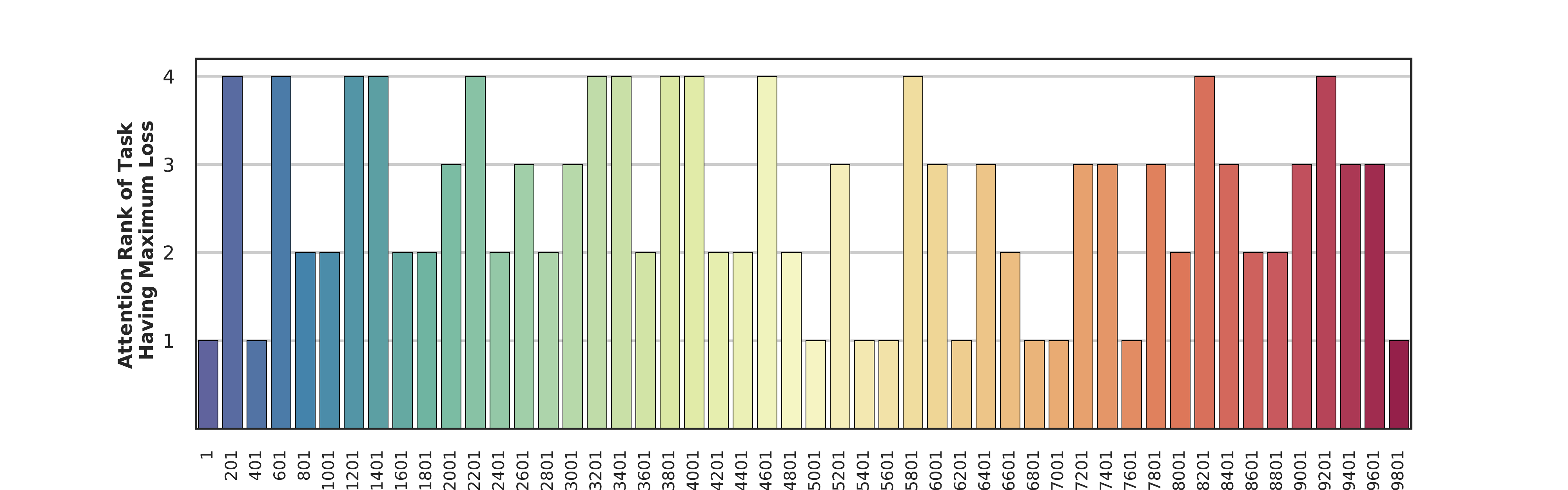}} 
\end{tabular}
 \caption{(a) Correlation of the components of meta-information (loss-ratio, accuracy, loss, gradient-norm) with attention vector (b) Rank (y-axis) of task having maximum loss as per the attention weight across training iterations (x-axis).}
\label{fig:correlation_analysis}
\end{figure}

We also analyze the mean Pearson correlation of each of the components (four tuple) of the meta-information with the attention vector across the training iterations. This is depicted in  Figure \ref{fig:correlation_analysis}(a) for TA-MAML in a 5-way 1-shot setting on the miniImageNet dataset.
 Overall, we observe that each of the individual meta-information has a  weak correlation with the attention weight, indicating that the TA module does not trivially follow any single component of meta-information.
We also observe that the loss-ratio and loss are positively correlated with attention vector, while accuracy and gradient norm are negatively correlated.
This suggests that the attention module, on average, assigns higher weights to less performing (larger loss) tasks over better performing tasks in a batch.
In human learning, this behavior is analogous to a student selecting topics lacking clarity to focus on for an exam preparation. However, to further verify that the loss is not the only criteria for providing weights to tasks, we plot the rank of the task incurring maximum loss in a batch, as per the attention weights across training iterations (Figure \ref{fig:correlation_analysis}(b)). A higher rank means more weight assigned to a task.  We observe that the TA module does not always assign maximum weight to the tasks that are having a high test loss. Thus, the TA module does not trivially learn to assign weights to the tasks based on some component of meta-information, but learns useful latent information from the provided meta-information to assign importance for the tasks in a batch.

\section{Summary and Future Work}
In this work we have shown that the batch wise episodic training regimen adopted by ML strategies can benefit from leveraging knowledge about the importance of tasks within a batch. Unlike prior approaches that assume uniform importance for each task in a batch, we propose task attention as a way to learn the importance of each task according to its alignment with the optimal meta-knowledge. We have  validated the effectiveness of task attention by augmenting it to popular initialization and parametric-optimization based ML strategies. To facilitate integration with the latter, we have introduced a batch wise training strategy for a parametric optimizer, that outperforms its previously sequential counterpart. 
We have demonstrated through few-shot learning experiments on miniImageNet and tieredImageNet datasets that augmenting task attention helps attain better  generalization to unseen tasks from the same distribution while requiring fewer iterations to converge. Visualization of the variation in the distribution of attention weights across the training iterations gives insights about the difference in the nature of meta-knowledge captured by initialization and optimization based ML strategies. We believe this provides an interesting future work direction for better understanding the meta-optimization landscape.

\section*{Acknowledgements}
The support and the resources provided by ‘PARAM Shivay Facility’ under the National Supercomputing Mission, Government of India at the Indian Institute of Technology, Varanasi and under Google Tensorflow Research award are gratefully acknowledged.
\bibliographystyle{splncs04}
\bibliography{samplepaper}

\end{document}